%% file: main.tex
\documentclass[letterpaper, 10 pt, conference]{ieeeconf}
\IEEEoverridecommandlockouts     
\usepackage{booktabs}
\usepackage{subfigure}
\usepackage{times}
\usepackage{helvet}
\usepackage{courier}
\usepackage{color}
\usepackage{amsmath}
\usepackage{amssymb}
\usepackage[ruled,vlined,linesnumbered]{algorithm2e}
\usepackage{url}
\usepackage{graphicx}
\usepackage{multirow}

\newtheorem{theorem}{Theorem}

\SetKwFor{Loop}{Loop}{}{EndLoop}
\title{\tda: Point Cloud Recognition Using\\Topological Data Analysis } 
\author{Anirban Ghosh, Iliya Kulbaka, Ian Dahlin, Ayan Dutta
\thanks{A. Ghosh, I. Dahlin,  and A. Dutta are with the School of Computing at the University of North Florida, USA.
Emails: {\tt\small \{anirban.ghosh, n01427009, n01487537, a.dutta\}@unf.edu}}%
}
\long\def\omitit#1{}

\newtheorem{problem}{Problem}

\newcommand{\tda}{\texttt{TopoRec}}

\begin{document}
\maketitle

\begin{abstract}
Point cloud-based object/place recognition remains a problem of interest in applications such as autonomous driving, scene reconstruction, and localization. Extracting a meaningful global descriptor from a query point cloud that can be matched with the descriptors of the database point clouds is a challenging problem. Furthermore, when the query point cloud is noisy or has been transformed (e.g., rotated), it adds to the complexity. To this end, we propose a novel methodology, named \tda, which utilizes Topological Data Analysis (TDA) for extracting local descriptors from a point cloud, thereby eliminating the need for resource-intensive GPU-based machine learning training. More specifically, we used the ATOL vectorization method to generate vectors for point clouds. To test the quality of the proposed \tda~technique, we have implemented it on multiple real-world (e.g., Oxford RobotCar, NCLT) and realistic (e.g., ShapeNet) point cloud datasets for large-scale place and object recognition, respectively. Unlike existing learning-based approaches such as PointNetVLAD and PCAN, our method does not require extensive training, making it easily adaptable to new environments. Despite this, it consistently outperforms both state-of-the-art learning-based and handcrafted baselines (e.g., M2DP, ScanContext) on standard benchmark datasets, demonstrating superior accuracy and strong generalization.
\end{abstract}

\input{intro}
\input{related}

\input{problem}

\input{tdacloud}

\input{exp}

\section{Conclusion and Future Work}
Large-scale point cloud recognition is a critical problem due to its practical relevance in applications such as loop closure detection for SLAM in robotics, global localization in autonomous vehicles, and more. Many state-of-the-art techniques rely on extensive training-based machine learning methods, where the trained model depends heavily on the training data and may not generalize well to unseen data. To address this, we propose a novel framework that analyzes the underlying topological properties of point clouds to generate a descriptor vector. This descriptor is then matched with the existing point clouds in a database, and the one with the highest similarity is returned as the solution in our proposed \tda~framework. When tested on several real-world datasets, including Oxford RobotCar and NUS (in-house), our lightweight \tda~technique significantly outperformed existing baselines, such as PointNetVlad and PCAN. The strong cross-day performance on the NCLT dataset, accurately localizing query scans within a small distance across different days and outperforming M2DP and ScanContext, demonstrates the robustness of our approach to temporal and environmental changes. These results highlight that topology-based, training-free place recognition methods can achieve strong performance, making them highly suitable for deployment in unseen and dynamic environments. Currently, \tda~does not require GPU support. In the future, we plan to incorporate small-scale GPUs to accelerate query processing, further enhancing the real-world applicability of \tda.

\bibliographystyle{ieeetr}
\bibliography{references}

\end{document}

%% file: intro.tex
\section{Introduction}

Scene understanding using 3D data is an active area of research in robotics and computer vision. Popular sub-problems include classification~\cite{qi2017pointnet}, segmentation~\cite{qi2017pointnet++}, and recognition~\cite{uy2018pointnetvlad}. In this paper, we study the 3D point cloud-based recognition problem. 
Point Cloud recognition is a fundamental component of autonomous driving, enabling vehicles to localize themselves accurately in dynamic, real-world environments. As autonomous systems navigate complex urban landscapes, precise mapping and real-time localization become crucial to ensure safe and efficient operation. LiDAR-based place recognition has emerged as a powerful solution for this task~\cite{zhang2024lidar}, offering high accuracy and robustness in challenging conditions where traditional vision-based methods often struggle, such as in low-light or adverse weather situations~\cite{uy2018pointnetvlad}. 

\begin{figure}[ht!]
    \centering
    \begin{tabular}{c|c}
\hspace*{-15pt}\includegraphics[width=0.65\linewidth]{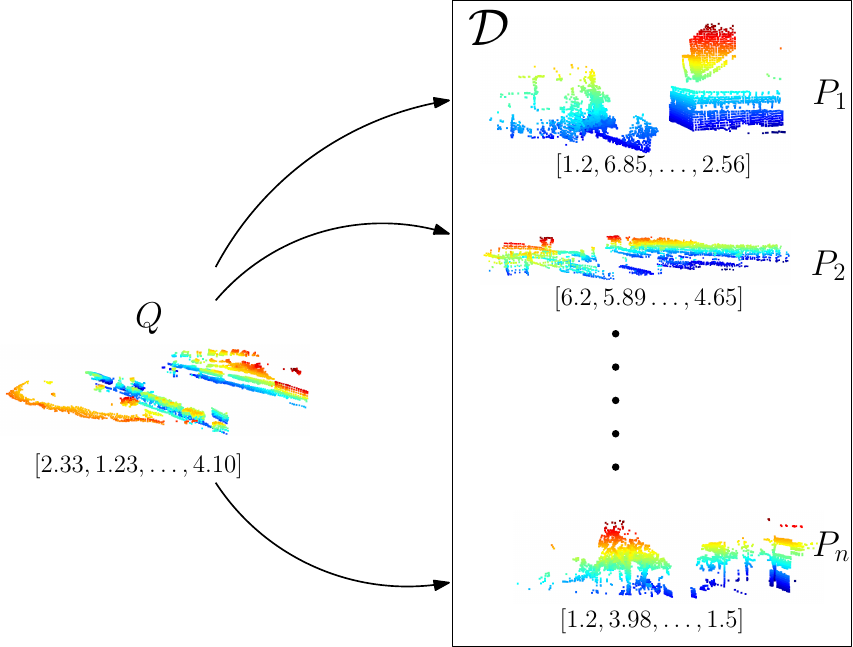} &\includegraphics[width=0.3\linewidth]{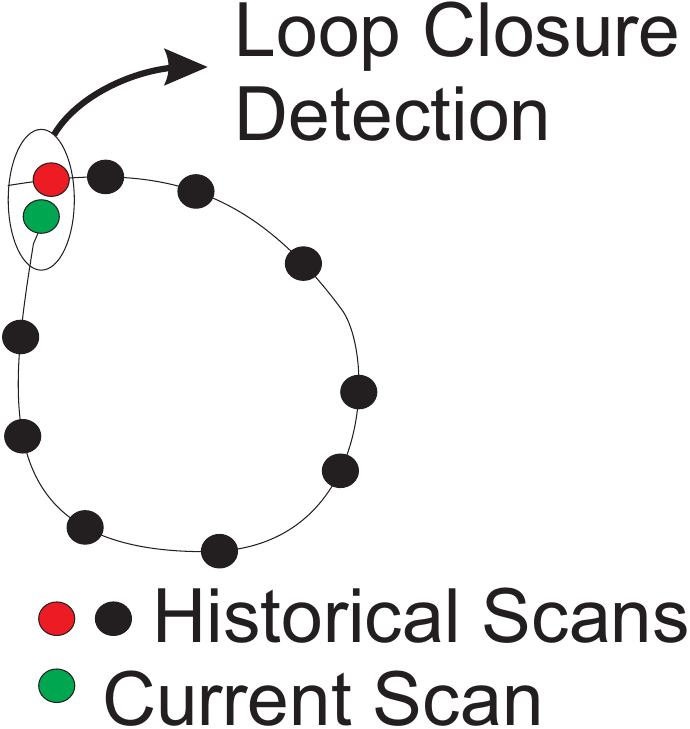}\\
    \end{tabular}
    \caption{a) A birds-eye view of \tda: a query LiDAR point cloud $Q$  
    is being matched against a database of point clouds $\mathcal{D}$ using a TDA-based descriptor vector; b) Motivation of place recognition: Loop Closure Detection (LCD) or ``where have I ever been" problem. The black curve represents the robot's trajectory, with solid circles showing the LiDAR scans collected over time. The solid green circle highlights the current scan, while the other circles correspond to previous scans. The green and red circles are nearby, and their scans share the highest similarity, completing the loop.}
    \vspace{-0.1in}
    \label{fig:intro}
\end{figure}

A particularly important challenge in autonomous driving and simultaneous localization and mapping (SLAM) is loop closure detection (LCD). Loop closure occurs when a robot revisits a previously mapped location, and recognizing this is essential for preventing drift in long-term navigation. This is also colloquially known as the ``where have I ever been" problem~\cite{zhang2024lidar,xiang2022delightlcd}. LiDAR-based place recognition significantly contributes to this problem by enabling reliable and efficient loop closure detection. By matching current/query point cloud observation with previously recorded scans, LiDAR systems can identify when a vehicle returns to a previously explored area, effectively ``closing the loop". This is critical for maintaining the consistency of the vehicle's map and preventing errors caused by drift, which can accumulate over time and degrade the accuracy of the SLAM system. See an illustration in Fig. \ref{fig:intro}(b).

Beyond improving localization accuracy, LiDAR-based place recognition supports other critical tasks such as path planning, obstacle avoidance, and traffic management, where GPS signals may be unreliable or unavailable.
One of the main challenges in large-scale point cloud-based place recognition is the high computational cost involved in processing and matching massive 3D datasets, especially in real-time applications. Additionally, variability in environmental conditions, such as changes in hardware specifications and dynamic objects, can lead to inconsistencies in point cloud data~\cite{uy2018pointnetvlad}, making accurate recognition and matching difficult. 
To this end, we propose a novel TDA-based approach for point cloud recognition. Although TDA has been leveraged to solve various point cloud applications~\cite{chazal2021introduction}, its efficacy in point cloud recognition for real-world datasets has not been previously studied.  Our presented framework, called \tda, uses a fast unsupervised vectorization method, ATOL~\cite{royer2021atol}.  The extracted ATOL vectors are robust to noise and transformations. We implemented \tda~in Python and tested on popular datasets for point cloud objects such as ShapeNet~\cite{shapenet2015} and Sydney Urban~\cite{de2013unsupervised}, as well as real-world datasets for large-scale place recognition such as those from Oxford~\cite{maddern20171} and NUS~\cite{uy2018pointnetvlad}, and NCLT~\cite{carlevaris2016university}. The results demonstrate the efficacy of the proposed technique. When compared against existing learning-based and handcrafted baselines such as PointNetVlad~\cite{uy2018pointnetvlad}, PCAN~\cite{zhang2019pcan}, M2DP~\cite{he2016m2dp}, and ScanContext~\cite{kim2018scan}, \tda~almost always outperformed them 17 times across various test cases, while getting outperformed by the baselines only twice, demonstrating its superiority in large-scale place recognition.

The main contributions of our work are:
\begin{itemize}
    \item To the best of our knowledge, \tda~is the first easy-to-use TDA-based approach that utilizes a TDA-based vectorization for large-scale real-world LiDAR point cloud recognition.
    \item Our proposed approach is lightweight (no extensive training) and does not require GPU support, unlike the state-of-the-art methods.
    \item We have tested the proposed \tda~framework on realistic as well as real-world datasets. Results show high recall values while outperforming popular baselines for the large-scale place recognition problem.
\end{itemize}

%% file: related.tex
\section{Related Work}
Point cloud recognition remains a popular research problem in the 3D computer vision and mobile robotics communities due to its importance in applications such as scene understanding/reconstruction and SLAM. A recent survey on various state-of-the-art techniques can be found in \cite{zhang2024lidar}. The authors in this survey also list various benchmark datasets - many of which we have selected in our paper for evaluation and comparative analysis. Many of the recent advancements are based on the foundational point cloud classification and segmentation models such as PointNet~\cite{qi2017pointnet} and PointNet++~\cite{qi2017pointnet++}. For example, one of the most popular large-scale place recognition studies, namely PointNetVlad~\cite{uy2018pointnetvlad} is a combination of PointNet and NetVLAD~\cite{arandjelovic2016netvlad}. We used the benchmark datasets provided by PointNetVlad and compared the performance of \tda~against it. Another example approach based on PointNet is PPFNet~\cite{deng2018ppfnet}, a permutation-invariant deep learning framework that accepts raw point clouds as inputs. The experiments were performed on the SUN-3D~\cite{xiao2013sun3d} dataset, which we also use in our experiments. A viewpoint-independent place recognition approach that relies on parallel semantic analysis of individual semantic attributes extracted from point clouds is proposed in PSE-Match~\cite{yin2021pse}. Another viewpoint-free technique is proposed in \cite{yin2021fusionvlad}, which includes an orientation-invariant as well as translation-insensitive feature
extraction modules.

Extracting feature descriptors (vectors) from point clouds is one of the main components used in the literature~\cite{zhang2024lidar}. Examples include~\cite{kim2021scan,sun2020dagc,liu2019seqlpd,yuan2024btc}. ScanContext++~\cite{kim2021scan} enhances robustness against rotational and lateral changes in urban place recognition. RING~\cite{lu2022one} extends the idea of global LiDAR descriptors such as ScanContext++ by leveraging a Radon sinogram representation, enabling not only place recognition but also relative orientation and translation estimation. The cross-day experiments with the NCLT dataset for large-scale place recognition in \cite{lu2022one} have motivated our experiment design. In \cite{sun2020dagc}, the authors propose a novel approach that integrates dual attention mechanisms and graph convolutional networks to improve place recognition accuracy with 3D point cloud data. The authors in \cite{liu2019seqlpd} utilize sequence matching and large-scale point cloud descriptors to improve the accuracy and efficiency of loop closure in autonomous driving. Delightlcd~\cite{xiang2022delightlcd} is a lightweight LCD technique that uses a dual-attention-based feature difference module in the deep network. These deep learning-based LCD techniques enhance localization accuracy for SLAM and autonomous driving. Unlike these state-of-the-art techniques, our proposed approach does not rely on extensive training or GPU computations.

%% file: problem.tex
\section{Problem Definition}

Our problem formulation in this paper follows from \cite{uy2018pointnetvlad,zhang2019pcan}. Let $\mathcal{D}$ be a database of $n$ point clouds $\{P_1,P_2,\ldots,P_n\}$ and $Q$ be a query point cloud (typically, $Q\notin \mathcal{D}$). Let $S(P)$ denote a vectorization function (a descriptor function) that generates a vector feature descriptor $v_P$, for any point cloud $P$. 
The point cloud recognition problem can be described as follows:

\begin{problem}
    Given a query point cloud $Q$, report a point cloud $P^* \in \mathcal{D}$, such that the structural difference between $Q$ and $P^*$ is the minimum possible.
    \label{def-prob}
\end{problem}

To solve Problem \ref{def-prob}, we pre-compute a feature descriptor $v_P:=S(P)$, for every $P \in \mathcal{D}$. When a query point cloud $Q$ arrives, its feature descriptor $v_Q:=S(Q)$ is computed. We return the point cloud $P \in \mathcal{D}$ as $P^*$ for which the  distance $d(\cdot)$ (depending on the descriptor function used) between $v_P,v_Q$ is the minimum:
\begin{equation}
    P^* = \arg \min_{P \in \mathcal{D}} d(v_P , v_Q)
    \label{eq}
\end{equation}

%% file: tdacloud.tex
\section{\tda: Our Proposed Methodology}
Topological data analysis (TDA) is the backbone of our proposed approach. With its roots in algebraic and computational topology, TDA has started to be widely used to analyze datasets from a geometric perspective in various research domains.  When appropriately extracted using computational topology algorithms, the geometric structures of datasets can reveal useful information about the datasets. The main focus of TDA is to provide sound
methods to infer, analyze, and exploit various
topological and geometric structures of data where the
data points are drawn from a metric space. \tda~uses persistent homology (PH) from TDA for real-world place recognition where the LiDAR scans are supplied as point clouds in the Euclidean $3$-space. Next, we briefly describe PH and the topological descriptors used in our work. For a comprehensive introduction to TDA and persistent homology, we refer the reader to~\cite{edelsbrunner2008persistent,chazal2021introduction}.

\begin{figure}[h]
    \centering
    \vspace{-10pt}
    \begin{tabular}{c c}
    \includegraphics[width=0.3\linewidth]{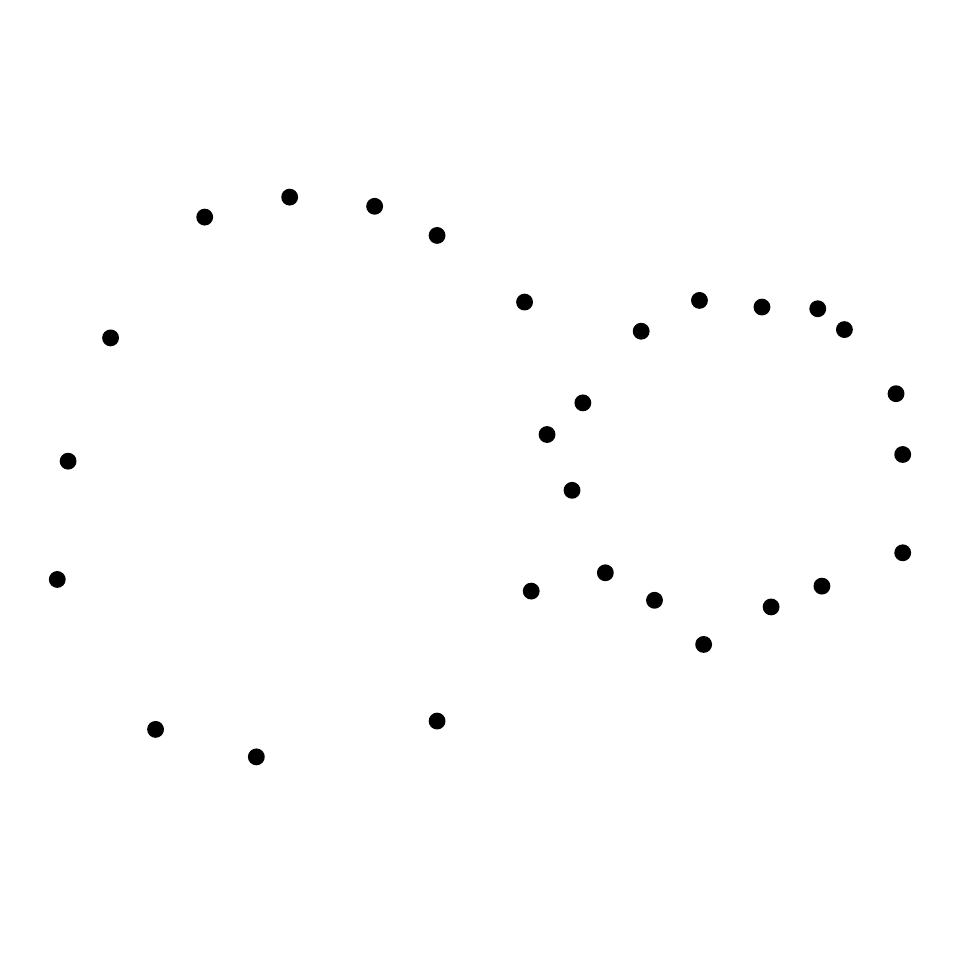}& 
    \includegraphics[trim=0 -1.5cm 0 0cm,width=0.3\linewidth]{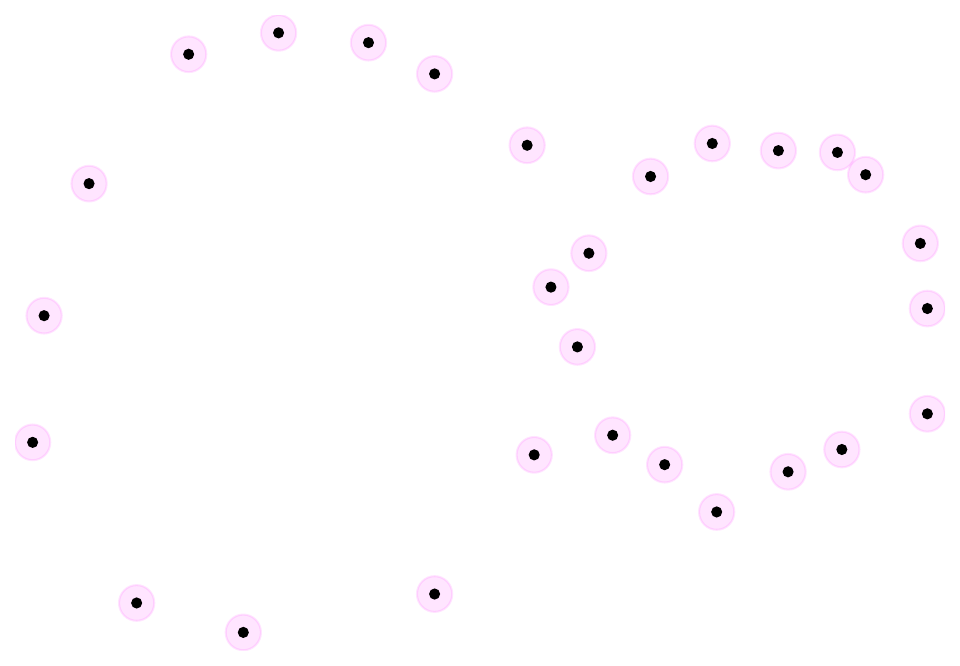} \vspace{-2pt}\\
    \vspace{-2pt}
    (a) &  (b) \\
    \includegraphics[width=0.3\linewidth]{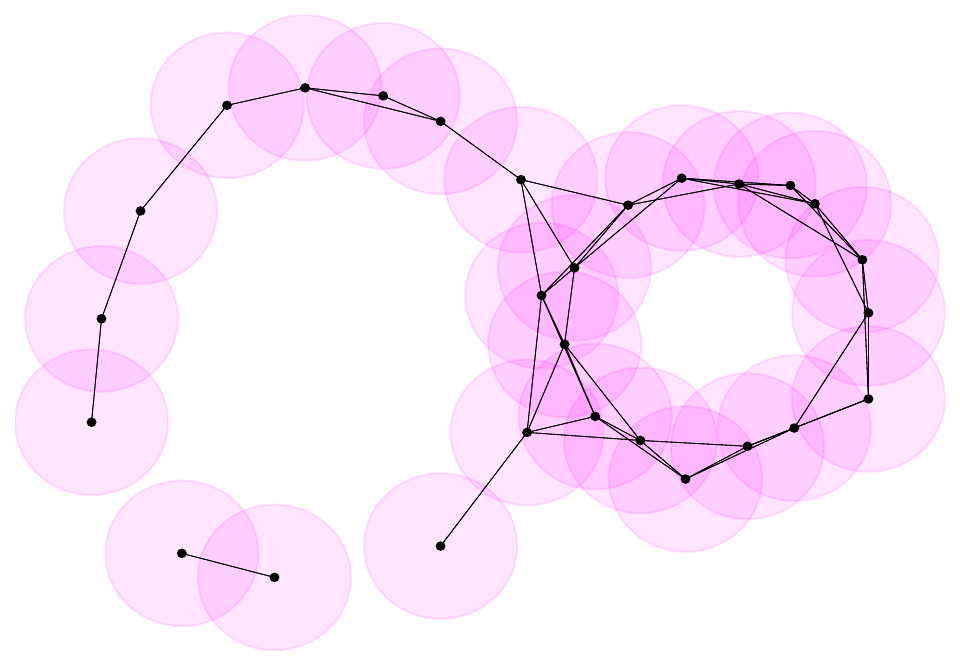}&
    \includegraphics[width=0.3\linewidth]{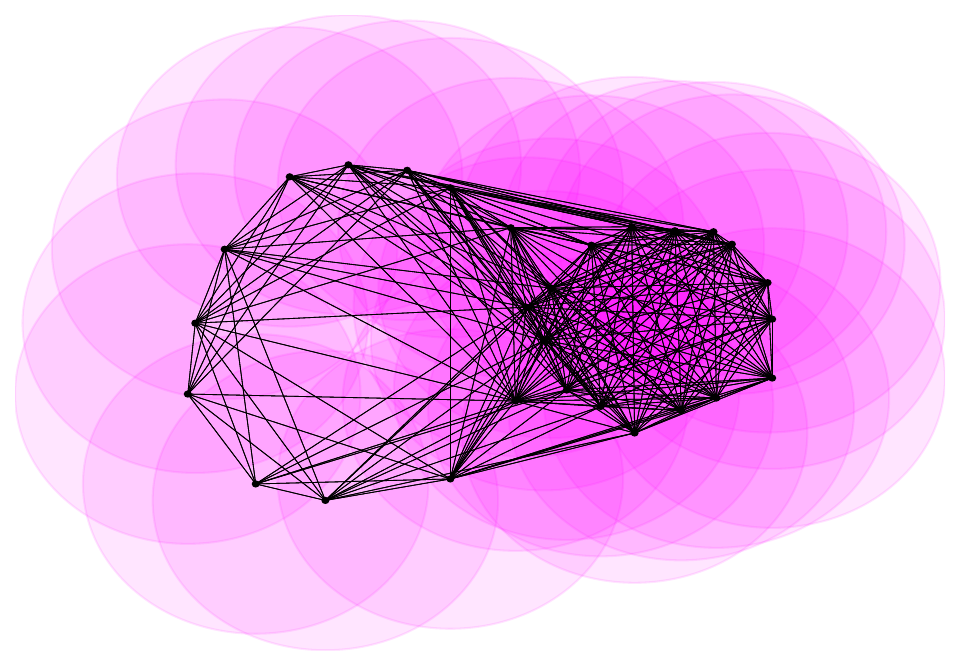} \\
    (c) &  (d) \\
     \hspace{-0.2in}\includegraphics[width=0.5\linewidth]{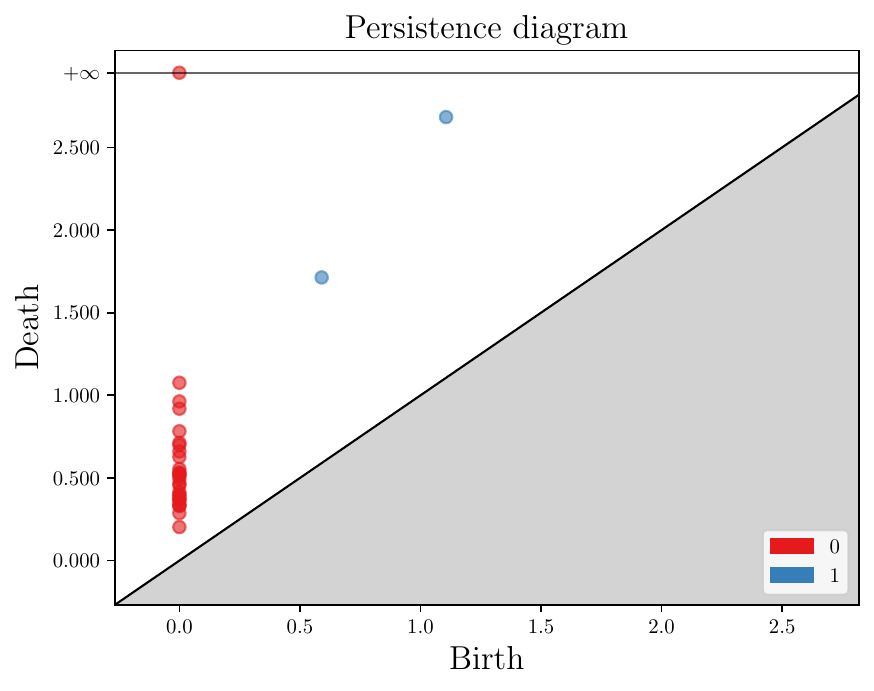} \hspace{-1.25in}&\includegraphics[width=0.55\linewidth]{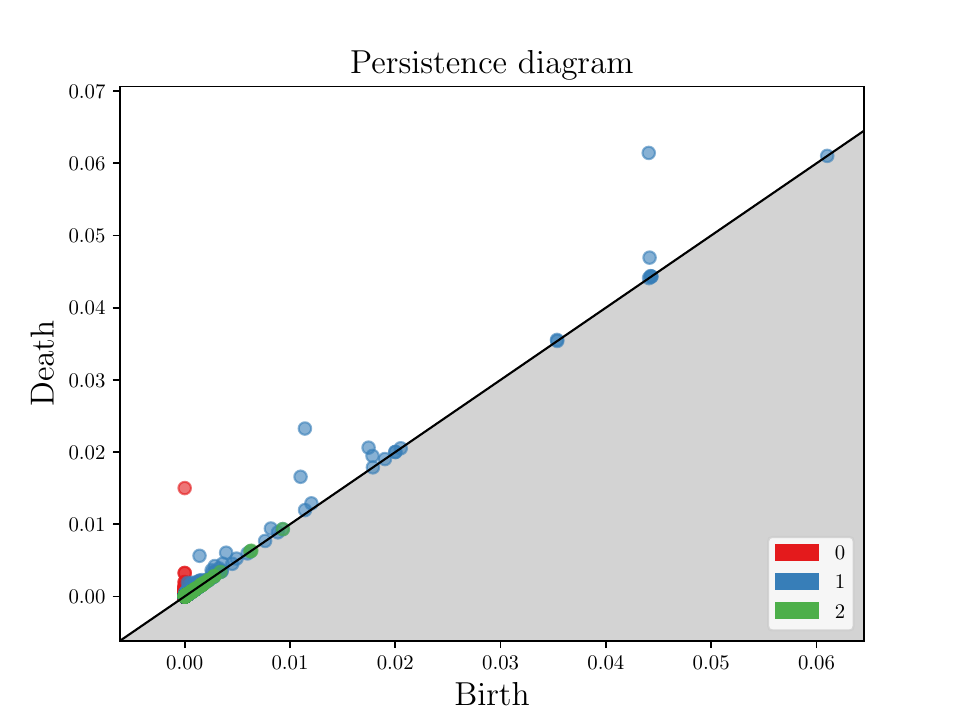} \\
       (e) & (f)
    \end{tabular}
      
\caption{Illustrating persistence homology on a $28$-element pointset (a). Some members of the family of nested simplicial complexes obtained by gradually increasing the radius of the disks centered at the points are shown in (b) -- (d). During the construction, an edge is placed between two points if their corresponding disks intersect.  The persistence diagram is shown in (e). The gray area is always empty since the death of a feature cannot occur before its birth. The persistence diagram for the 3D query point cloud $Q$ in Fig. \ref{fig:intro}(a) is shown in (f). The red, blue, and green dots correspond to the homology dimensions $0,1,2$, respectively. }
\label{fig:homology-demo}
\end{figure}    

PH can detect interesting topological features, such as connected components, holes, and cavities, corresponding to homology dimensions $0,1,2$, respectively. In PH, similar-looking point clouds tend to have similar kinds of topological features. PH encodes topological features of nested families of simplicial complexes (geometric graphs where the vertices are points of the point cloud and the edges capture closeness), formed by increasing the common radius $r$ (set to $0$ at the start) of every ball placed centered at each point in the given point cloud. If two balls intersect, an edge is placed between their corresponding points. Therefore, we obtain a simplicial complex for every value of $r$, resulting in a sequence of growing simplicial complexes where every complex is a subgraph of its successor in the sequence, known as a filtration. The sequence is analyzed using a persistence descriptor that keeps track of the features (connected components, holes, and cavities) with increasing $r$. If a feature appears at radius $r=r_i$ and dies when at $r=r_j$, where $r_j> r_i$, the pair $(r_i,r_j)$ forms a birth-death pair for the feature.  The list of birth-death pairs, known as the persistence of the point clouds, is a topological descriptor for a point cloud. Similar-looking point clouds are expected to have almost identical lists of birth-death pairs. The birth-death pairs (a set of $(r_i,r_j)$ points) can be visualized using a persistence diagram, a scatter plot of the birth-death pairs for different homology dimensions ($0,1,2$ for 3D point clouds). 
For the sake of brevity, we provide a toy example in Fig.~\ref{fig:homology-demo} to explain the idea of persistence and the persistence diagram.  

In our experiments, we primarily used persistence in homology dimension $2$. However, real-world LiDAR point clouds do not always admit enough birth-death pairs in dimension $2$ due to an insufficient number of cavities in the point clouds. In that case, we obtain the persistence in homology dimension $1$. If we fail again, we use dimension $0$, which is guaranteed to have at least $|p|$ birth-death pairs, where $p$ is the point cloud under consideration.

\begin{algorithm}
\small{
\caption{\texttt{\textbf{\tda}}}
\label{fig:alg}
\KwData{$Q$, $\mathcal{D}=\{P_1,P_2,\ldots,P_n\}$, $b$}
\KwResult{$P^* \in \mathcal{D}$}
\textbf{Onetime pre-processing: generating a suitable persistence $\forall P\in \mathcal{D}$.} Generate persistence $\texttt{\textbf{pers}}_{P}$ (a list of birth-death pairs) for every $P \in \mathcal{D}$ in homology dimension $2$. If the $\texttt{\textbf{pers}}_{P}$ has less than $2$ birth-death pairs, generate the persistence in homology dimension $1$ and use it as $\texttt{\textbf{pers}}_{P}$. Use the persistence in dimension $0$ as $\texttt{\textbf{pers}}_{P}$, if dimension $1$ fails to generate at least $2$ birth-death pairs\;
\textbf{Onetime pre-processing: generating a vector $\forall P \in \mathcal{D}$.} Generate a $b$-length ATOL descriptor vector $v_P$ using $\texttt{\textbf{pers}}_{P}$, for every $P \in \mathcal{D}$, where $v_P=[a_{P1},a_{P2},\ldots,a_{Pb}]$\;
Generate the query $b$-length ATOL vector $v_Q=[c_1,c_2,\ldots, c_b]$ for $Q$ using its persistence, obtained using the same approach used for the point clouds in $\mathcal{D}$\;
Return the point cloud $P^* \in \mathcal{D}$  such that $P^* = \arg \min_{P \in \mathcal{D}} ||v_P - v_Q||$, where $||\cdot||$ denotes the Euclidean distance in $b$-space\;
}
\end{algorithm}

As a part of pre-processing, for each $P \in \mathcal{D}$, we obtain a $b$-length vector $v_P = [a_{P1}, a_{P2},\ldots,a_{Pb}]$ from its persistence using the unsupervised ATOL (Automatic Topologically-Oriented Learning) vectorization method introduced in~\cite{royer2021atol}. The optimal length $b$ that gives the best recall value typically depends on the type of the dataset being considered and can vary from one dataset to another. When a query point cloud $Q$ arrives, we compute its persistence akin to the pre-processing step and find its corresponding ATOL descriptor vector $v_Q=[c_1,c_2,\ldots,c_b]$.  The point cloud $P^*\in D$ is returned as the closest match for $Q$, such that the Euclidean distance between $v_{P^*}$ and $v_Q$ is the lowest among all $P \in \mathcal{D}$; refer to Eq.~\ref{eq}. If the ATOL vectors $v_{P_1},v_{P_2},\ldots,v_{P_n}$ are pre-processed in a $k$d-tree, $P^*$ can be reported in $O(\log n)$ time for a single query $Q$, for a fixed value of $b$. 

A pseudocode for our approach is presented in Algorithm 1. \texttt{\tda} does not require extensive training time to set up ATOL for generating vectors for the point clouds in $\mathcal{D}$ and $Q$. Consequently, our approach is lightweight compared to other machine learning-based approaches in the literature. In Theorem~\ref{thm:toporec}, we argue that \tda~is permutation, rotation, and translation invariant, meaning even if the points are shuffled and/or the point clouds are rotated or translated,  \tda~will always return the same matched point cloud $P^*$, irrespective of the modification(s) applied.

\begin{theorem}
\tda~is permutation, rotation, and translation invariant.
\label{thm:toporec}
\end{theorem}

\begin{proof}
Let $P$ and $P'$ be two point clouds such that $P'$ has been obtained from $P$ by permuting the points in $P$, rotating $P$, and/or translating $P$ in a specific direction. We observe that after applying one or more of such modifications on $P$, the topological features in every homology dimension remain the same in $P'$ since all pairwise distances between the points remain unchanged in $P'$. Consequently, the persistences of $P$ and $P'$ are identical, as the birth-death pairs in persistences are independent of the actual coordinates of the points. Two identical persistences admit the same ATOL vector. Consequently, irrespective of the modifications applied to the point clouds in $\mathcal{D}\cup \{Q\}$,~\tda~will always return the same $P^* \in \mathcal{D}$. This establishes our claim.
\end{proof}

 However, if the pairwise point pair distance is different in $P'$, the persistences of $P$ and $P'$ are not guaranteed to be the same. For instance, if $P'$ is obtained from $P$ by scaling, the ATOL vector of $P$ will likely be different from that of $P$. The above claims are also observed in our experiments in Section~\ref{sec:exp}.

%% file: exp.tex
\section{Experiments and Results}
\label{sec:exp}
\subsection{Settings}

A desktop with an {i9-12900K} processor and $32$-GB of main memory was deployed for the experiments. The `gudhi' library, popular for TDA in Python, was used to generate persistences (using {gudhi.AlphaComplex}) and ATOL vectors. The Open3D library was used to create visualizations of point clouds and process the database and query point clouds. The ATOL vector generator needs a quantizer and a contrast function. In our experiments, we chose {MiniBatchKMeans} from {scikit-learn} for speed and {laplacian} as the contrast function (supplied by the gudhi library) for high recall. 

\subsection{Datasets}
\textbf{Point Cloud Object Recognition Datasets. }
The ShapeNet dataset~\cite{shapenet2015} is a large-scale collection of 3D models designed for research in computer vision and robotics. It contains over 3 million 3D models across 55 object categories, with a rich set of annotations such as object parts, poses, and semantics. We have used $800$ point clouds randomly selected from eight categories ($100$ each) as our point cloud database. The Sydney Urban Point Cloud dataset~\cite{de2013unsupervised} consists of high-resolution LiDAR point cloud data captured from various urban environments in Sydney, Australia. It has $13$ classes and $588$ point clouds. 

For the KITTI-360~\cite{Liao2022PAMI} dataset, we used the fused LiDAR scans from Velodyne HDL-64E and a SICK LMS-200. Specifically, we used trajectories $[0,2,3,4,5,6,7,9,10]$. From each of these trajectories, we selected 20\% point clouds as queries. For the SUN-3D dataset~\cite{xiao2013sun3d}, we selected $1000$ random depth images, which were then converted to 3D point clouds by the authors' provided code\footnote{https://sun3d.cs.princeton.edu/}. Like KITTI-360, we used 20\% point clouds as queries, and the corresponding results are reported next. For these two datasets, the database and the query clouds were uniformly downsampled to approximately $10,000$ points for our experiments. 

\noindent
\textbf{Oxford and NUS Datasets. }The Oxford RobotCar and NUS (in-house) datasets are popular in the literature on place recognition. We chose two prior studies that used these datasets, PointNetVlad and PCAN, as baselines for comparative analysis. Oxford and NUS LiDAR scan datasets were collected in the real world using SICK LMS-151 and Velodyne HDL-64, respectively. All point clouds in these datasets were preprocessed by removing the ground planes and downsampling to $4096$ points. The point coordinates were then shifted and rescaled to have a zero mean and fall within [-1,1]. Following \cite{uy2018pointnetvlad}, to assess \tda~'s ability to generalize across various runs of mapping, we query a submap from a test reference map against a database built from all submaps of another reference map collected during a separate run of the same environment. If the retrieved point cloud scan $P^*$ is within 25m of the ground truth (revisit criteria), then we call it a success. We used the baseline test setting from \cite{uy2018pointnetvlad}. For further information on the Oxford and NUS datasets, refer to \cite{uy2018pointnetvlad}.

\noindent
\textbf{NCLT Dataset. }We also verified the strength of \tda~ in a large-scale place recognition, where the observed and query LiDAR point cloud datasets were collected on different days. For this, we have used the large-scale NCLT dataset~\cite{carlevaris2016university}. Collected by a Segway robot at the University of Michigan North Campus, the NCLT dataset is a comprehensive, long-term, and large-scale resource. It features 27 sessions, recorded approximately every two weeks over 15 months, spanning all four seasons. The NCLT dataset differs from Oxford and NUS in that the number of points per LiDAR scan is not fixed and varies considerably across scans, generally being larger than that of Oxford and NUS. The revisit distance criteria vary between $\{5, 10, 20, 50\}$m. In our evaluation, we designated the trajectory dated $X$ as the map database $\mathcal{D}$, with the trajectories dated $Y$ functioning as query data. There are four such combinations used in our experiments as listed below.

\begin{table}[ht!]
    \centering
    \begin{tabular}{ccc}
         & Map (M) & Query (Q) \\
         \hline
        Case 1 & 2012-02-04 & 2012-03-17\\
        Case 2 & 2012-03-17 & 2012-02-04\\
        Case 3 & 2013-02-23 & 2013-04-05\\
        Case 4 & 2013-04-05 & 2013-02-23\\
        \hline
    \end{tabular}
    \caption{Used NCLT datasets for place recognition with cross-day map database and query data.}
    \label{tab:NCLT_MQ}
\end{table}

For performance evaluation, we have mainly looked into two metrics - Recall @$N$ (the percentage of queries for which at least one of the top $N$ retrieved results is a correct match to the ground truth) and Recall-1\% (the percentage of queries for which the top 1\% of retrieved results include a correct match to the ground truth). The average recall values (in \%) yielded by this evaluation are presented in the next section. The number of queries used in our evaluations is listed in Table \ref{tab:query_count}. We also report the runtime of \tda~to process a query and retrieve the recognized point clouds.

\subsection{Results}
\subsubsection{Point Cloud Object Recognition}
First, we applied our proposed \tda~framework on a realistic point cloud dataset, ShapeNet. The Recall @1 and Recall-1\% values (in \%) are shown in Table \ref{tab:shapenet-800}. Given that this dataset is not collected in the real world, we introduced artificial noise to the queries to make it more realistic. The list of added noise types can be found in Table \ref{tab:shapenet-800}. We found that when the query $Q$ was not perturbed, recall values were $100\%$, i.e., \tda~worked perfectly. On the other hand, when noise is added, \tda~performs still yields almost $100\%$ Recall-1\% values. Recall @1 values are the lowest when $Q$ is down-scaled by $5\%$ (scaling is done by multiplying the coordinates of every point by $1 \pm 0.05$). When the points in $Q$ are jittered by  $5\%$, i.e., a white Gaussian noise ($\mu:0, \sigma:10^{-3}$) is added to the locations of those points, \tda~still performs almost perfectly. If the points in a point cloud change their locations, the structure of the point cloud changes, adding a challenge to the operation of \tda. However, it still performs strongly in the Recall-1\% metric and achieves $100\%$ accuracy. A similar performance trend was observed for the Sydney Urban dataset (see Table \ref{tab:sydney}). However, for the jitter noise, the performance was found to be slightly worse than ShapeNet.

\begin{table*}[ht!]
    \centering
    \begin{tabular}{p{1cm}p{1.2cm}p{1cm}p{0.75cm}p{1cm}p{1.2cm}p{1cm}p{1.2cm}p{1.2cm}p{1.2cm}p{1.2cm}}
    \toprule
        Dataset & ShapeNet & Sydney Urban & Oxford & NUS & SUN-3D & KITTI-360 & NCLT (Case 1) & NCLT (Case 2) & NCLT (Case 3) & NCLT (Case 4)\\
        \hline
        Queries
         & $104$
         & $117$
         & $3030$
         & 1751
         & 200
         & 63
         & 20981
         & 19983
         & 16493
         & 19137 \\
         \bottomrule
    \end{tabular}
    \caption{Number of queries used for evaluation in each dataset.}
    \label{tab:query_count}
\end{table*}

\begin{table}[h]
    \centering
    \begin{tabular}{|c||c|c|c|c|}
    \hline
     {Modification} & {Recall @1} & {Recall $1\%$}          \\ \hline \hline 
    Original    &   $100$       &   $100$               \\ \hline

    $5\%$-jitter ($\mu:0,\sigma:10^{-3}$)   &   $97.11$     &  $100$                \\ \hline

    $+5\%$-scaling    &     $75$     &    $99.04$               \\ \hline
    $-5\%$-scaling     &    $56.73$      &    $99.04$                \\ \hline

    [$0.2, 0.2, 0.1$]-translation    &    $100$      &  $100$               \\ \hline
    
    $45^{\circ}$-rotation    &    $100$      &   $100$                \\ \hline
    $90^{\circ}$-rotation    &     $100$     &   $100$                 \\ \hline
    $135^{\circ}$-rotation    &    $100$      &    $100$               \\ \hline

    \end{tabular}
    \caption{Recall results (\%) for the \texttt{Shapenet} dataset. $b=10$}
    \label{tab:shapenet-800}
\end{table}

\begin{table}[h]
    \centering
    \begin{tabular}{|c||c|c|c|c|}
    \hline
     {Modification} & {Recall @1} & {Recall $1\%$}        \\ \hline \hline 
    Original    &   $100$       &   $100$                 \\ \hline

    $5\%$-jitter ($\mu:0,\sigma:10^{-3}$)   &   $94.02$     &  $98.29$                \\ \hline

    $+5\%$-scaling    &     $72.65$     &    $98.29$                \\ \hline
    $-5\%$-scaling     &    $70.94$      &    $97.44$                \\ \hline

    [$0.2, 0.2, 0.1$]-translation    &    $100$      &  $100$                 \\ \hline
    $45^{\circ}$-rotation    &    $100$      &   $100$                \\ \hline
    $90^{\circ}$-rotation    &     $100$     &   $100$                 \\ \hline
    $135^{\circ}$-rotation    &    $100$      &    $100$                \\ \hline

    \end{tabular}
    \caption{Recall results (\%) for the \texttt{Sydney Urban} dataset. $b=10$}
    \label{tab:sydney}
\end{table}

\begin{table}[h]
    \centering
    \begin{tabular}{|c||c|c|c|c|}
    \hline
     {Modification} & {Recall @1} & {Recall $1\%$}        \\ \hline \hline 
    Original    &   $100$       &   $100$                 \\ \hline

    $5\%$-jitter ($\mu:0,\sigma:10^{-3}$)   &   $100$     &  $100$                \\ \hline

    $+5\%$-scaling    &     $83.5$     &    $100$                \\ \hline
    $-5\%$-scaling     &    $79.5$      &    $100$                \\ \hline

    [$0.2, 0.2, 0.1$]-translation    &    $100$      &  $100$                 \\ \hline
    $45^{\circ}$-rotation    &    $100$      &   $100$                \\ \hline
    $90^{\circ}$-rotation    &     $100$     &   $100$                 \\ \hline
    $135^{\circ}$-rotation    &    $100$      &    $100$                \\ \hline

    \end{tabular}
    \caption{Recall results (\%) for the \texttt{SUN-3D} dataset. $b=10$}
    \label{tab:sun3d}
\end{table}

\begin{table}[h]
    \centering
    \begin{tabular}{|c||c|c|c|c|}
    \hline
     {Modification} & {Recall @1} & {Recall $1\%$}        \\ \hline \hline 
    Original    &   $100$       &   $100$                 \\ \hline
    $5\%$-jitter ($\mu:0,\sigma:10^{-3}$)   &   $100$     &  $100$                \\ \hline
   
    $+5\%$-scaling    &     $68.25$     &    $68.25$                \\ \hline
    $-5\%$-scaling     &    $71.43$      &    $71.43$                \\ \hline

    [$0.2, 0.2, 0.1$]-translation    &    $100$      &  $100$                 \\ \hline
    
    $45^{\circ}$-rotation    &    $100$      &   $100$                \\ \hline
    $90^{\circ}$-rotation    &     $100$     &   $100$                 \\ \hline
    $135^{\circ}$-rotation    &    $100$      &    $100$                \\ \hline

    \end{tabular}
    \caption{Recall results (\%) for the \texttt{KITTI-360} dataset. $b=10$}
    \label{tab:kitti360}
\end{table}

\begin{figure}[h]
    \centering
    \begin{tabular}{c c}
    \fbox{\includegraphics[width=0.4\linewidth]{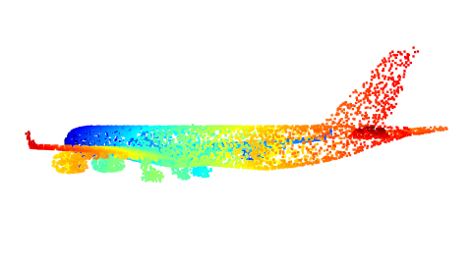}} &\hspace{-0.2in}
    {\setlength{\fboxrule}{2pt}\fcolorbox{green}{white}{\includegraphics[width=0.4\linewidth]{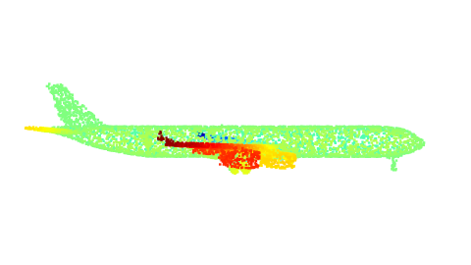}}} \\
    (a) &  (b) \\
    
    \fbox{\includegraphics[width=0.4\linewidth]{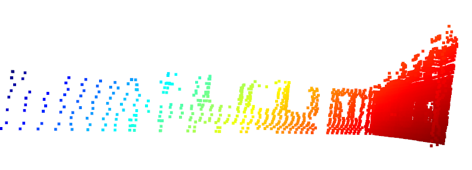}} &\hspace{-0.2in}
    {\setlength{\fboxrule}{2pt}\fcolorbox{green}{white}{\includegraphics[width=0.4\linewidth]{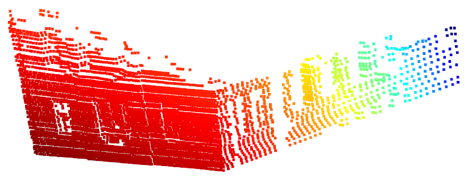}}} \\
    (c) &  (d) \\
    \fbox{\includegraphics[width=0.4\linewidth]{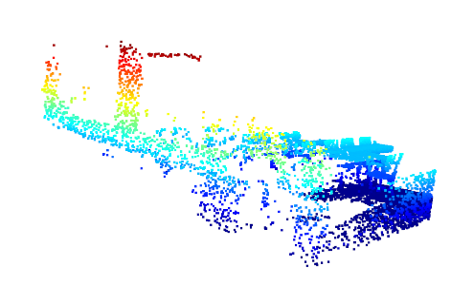}} &\hspace{-0.15in}
    {\setlength{\fboxrule}{2pt}\fcolorbox{green}{white}{\includegraphics[width=0.4\linewidth]{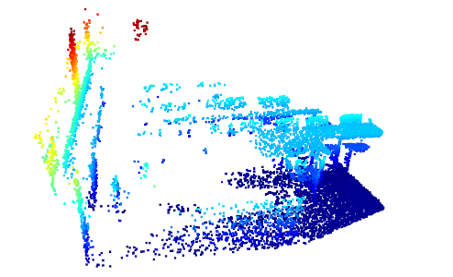}}} \\
    (e) &  (f) \\
    \fbox{\includegraphics[width=0.4\linewidth]{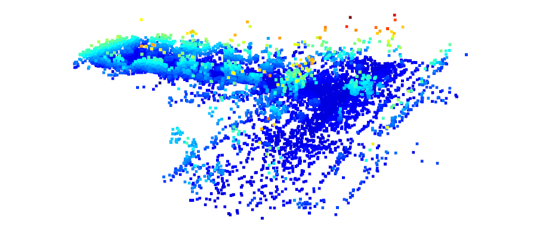}} &\hspace{-0.15in}
    {\setlength{\fboxrule}{2pt}
\fcolorbox{green}{white}{\includegraphics[width=0.4\linewidth]{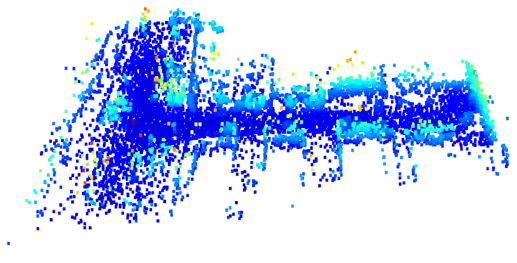}}} \\
    (g) &  (h) \\
    \end{tabular}
      
\caption{The query on the left was a 135$^\circ$ rotated version of a database point cloud (retrieved on the right) from Shapenet  (a, b), Sydney Urban (c, d), SUN-3D  (e, f), and KITTI-360 (g, h).}
\label{fig:queryandmatch2}
\end{figure}

Next, we performed the same experiments on large-scale datasets used specifically for place recognition, i.e., SUN-3D and KITTI-360. When \tda~was tested with the SUN-3D dataset, the results were stronger than both the ShapeNet and Sydney Urban datasets. Only for the scaling transformation, the Recall @1 metric goes to around 80\% whereas the Recall-1\% metric is always 100\% (see Table \ref{tab:sun3d}). We again noticed that when the query point cloud was scaled up or down, \tda~struggled, it could still get around 70\%  on average in those two scale noise cases. Interestingly, in the KITTI-360 dataset, the top 1\% was represented by just one entry in each of the trajectory folders. As a result, the recall values in Table \ref{tab:kitti360} are identical. Overall, we can see a very high level of accuracy under various perturbations, which showcases the robustness of \tda. Examples of successful retrievals for these four datasets are shown in Fig.~\ref{fig:queryandmatch2}.

The average query times for Shapenet, Sydney Urban, SUN-3D, and KITTI-360 were 500 ms, 30 ms, 440 ms, and 60 ms, respectively. The differences in the run times are mainly due to the variation in persistence generation time, as 99\% of the query time is spent computing the persistence of the query point cloud. 

\begin{table}[ht!]
    \centering
    \begin{tabular}{ccccc}
    \toprule
        Method \textbackslash ~Dataset & Oxford & NUS-U & NUS-R & NUS-B\\ \hline
        PointNetVLAD~\cite{uy2018pointnetvlad}& 80.30 & 72.60 & 60.3 & 65.30\\
        PCAN~\cite{zhang2019pcan}& 83.8 & 79.10 & 71.20 & 66.80\\
        \hline
        M2DP~\cite{he2016m2dp}& 95.21 &  78.09 & \textbf{83.06} & 73.36\\
        ScanContext~\cite{kim2018scan}& 92.01 & 79.12 & 76.88 & \textbf{75.08}\\
         \hline
        \tda~(Ours)& $\mathbf{95.60}$ & $\mathbf{82.90}$ & ${81.70}$ & ${73.20}$\\
        \bottomrule 
    \end{tabular}
    \caption{Average Recall-$1\%$ Values (\%). Oxford: $b=50$, NUS-U/R/B: $b=1000$}
    \label{tab:recall1p_comp}
\end{table}

 \begin{figure}[h]
    \centering
    \setlength{\fboxsep}{0pt}
    \begin{tabular}{c c}
    \fbox{\includegraphics[width=0.5\linewidth]{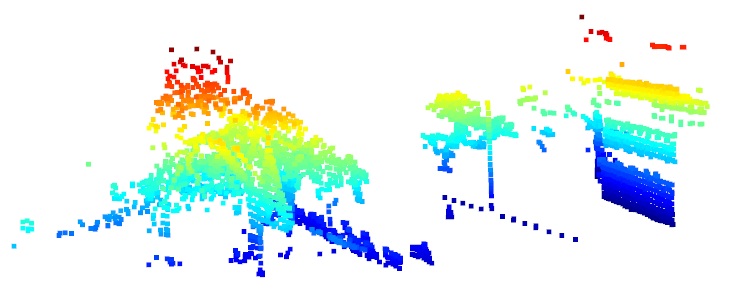}}&
    \hspace{-0.15in}{\setlength{\fboxrule}{2pt}\fcolorbox{green}{white}{\includegraphics[width=0.5\linewidth]{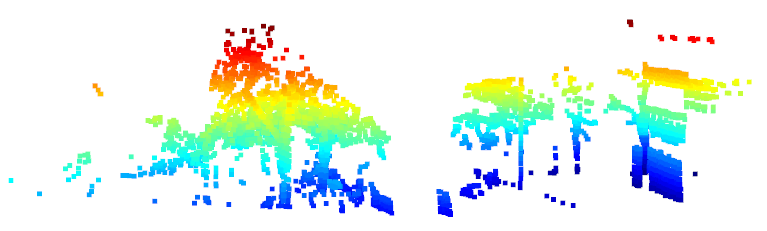}}} \\
    (a) &  (b) \\
    \fbox{\includegraphics[width=0.5\linewidth]{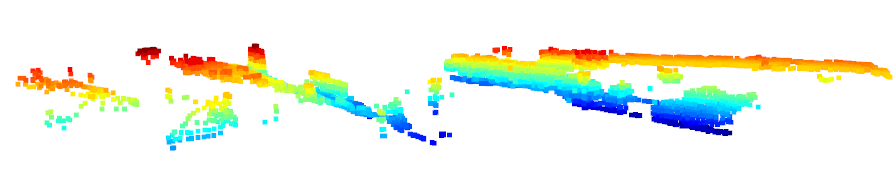}}&
    \hspace{-0.15in}{\setlength{\fboxrule}{2pt}\fcolorbox{green}{white}{\includegraphics[width=0.5\linewidth]{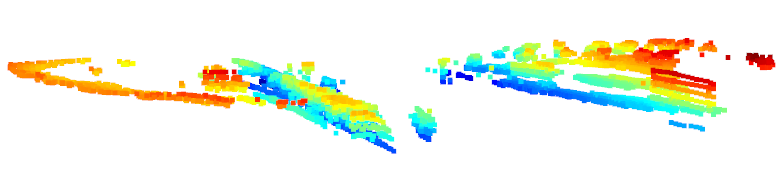}}} \\
    (c) &  (d) \\
    \fbox{\includegraphics[width=0.5\linewidth]{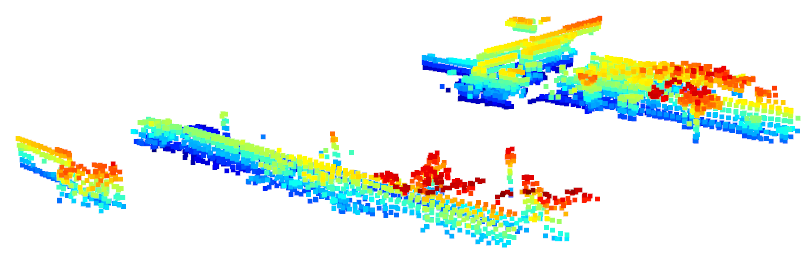}}&
    \hspace{-0.15in}{\setlength{\fboxrule}{2pt}\fcolorbox{green}{white}{\includegraphics[width=0.5\linewidth]{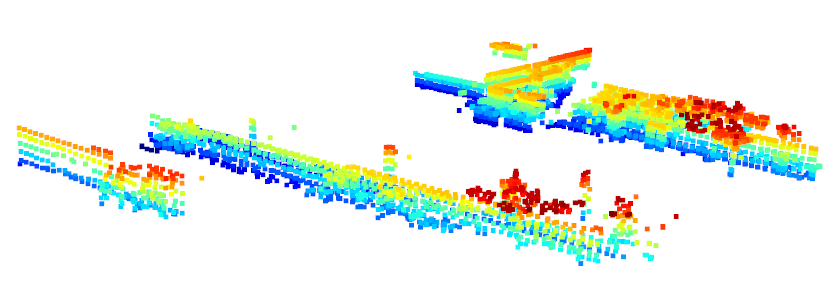}}} \\
    (e) &  (f) \\
    \fbox{\includegraphics[width=0.5\linewidth]{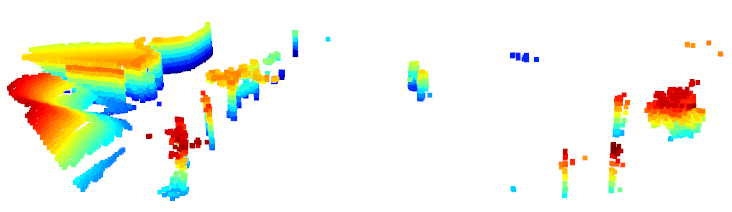}}&
    \hspace{-0.15in} {\setlength{\fboxrule}{2pt}\fcolorbox{green}{white}{\includegraphics[width=0.5\linewidth]{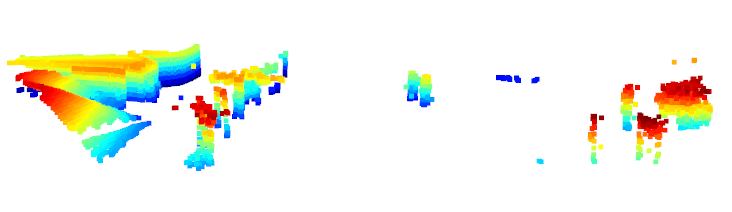}}}\\
    (g) &  (h) \\
    \end{tabular}
      
\caption{A query (left) and its corresponding matched point cloud (right) from Oxford (a, b), NUS-U (c, d), NUS-R  (e, f), and NUS-B  (g, h). 
}
\label{fig:queryandmatch_oxford}
\vspace*{-2pt}
\end{figure}    

 \begin{figure}[h]
    \centering
    \begin{tabular}{c c}
    \fbox{\includegraphics[width=0.5\linewidth]{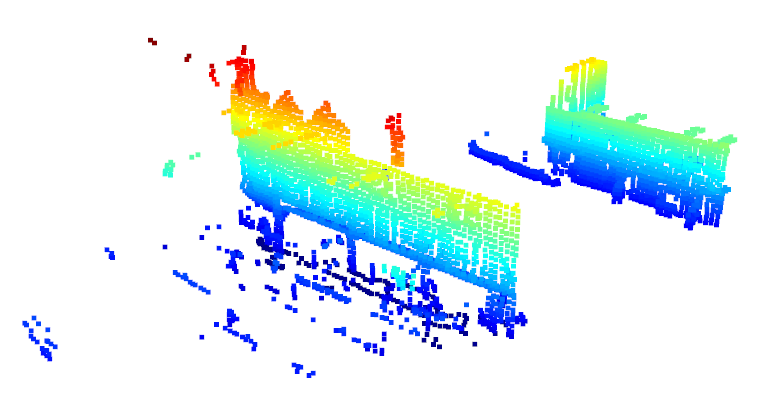}}&
    \hspace{-0.15in}{\setlength{\fboxrule}{2pt}\fcolorbox{red}{white}{\includegraphics[width=0.5\linewidth]{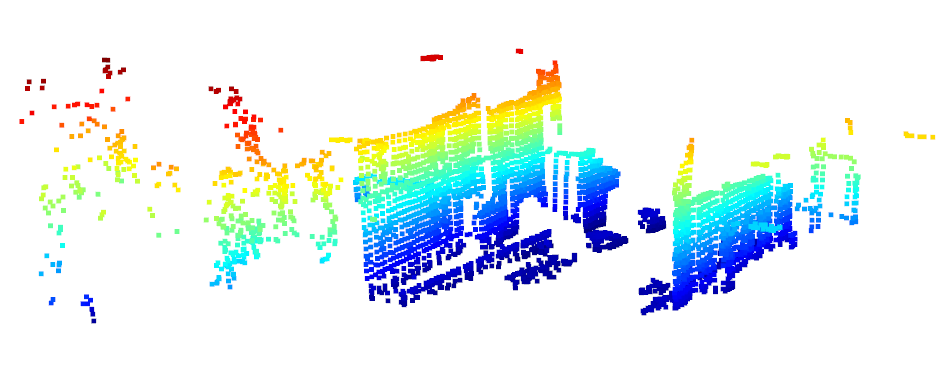}}}\\
    (a) &  (b) \\
    \fbox{\includegraphics[width=0.5\linewidth]{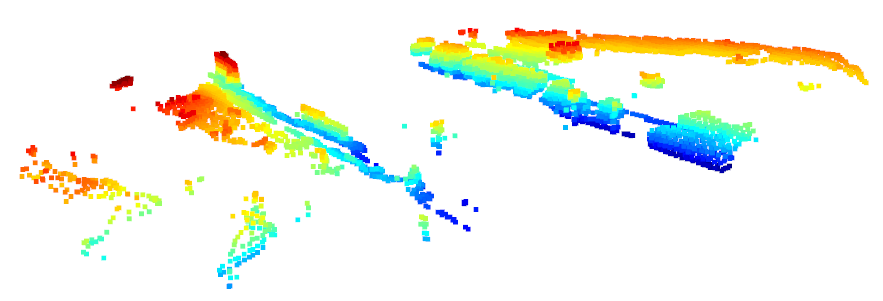}}&
    \hspace{-0.15in}{\setlength{\fboxrule}{2pt}\fcolorbox{red}{white}{\includegraphics[width=0.5\linewidth]{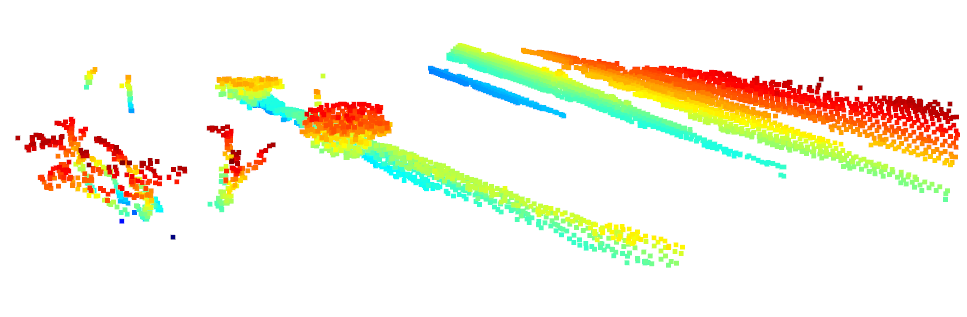}}} \\
    (c) &  (d) \\
    \fbox{\includegraphics[width=0.5\linewidth]{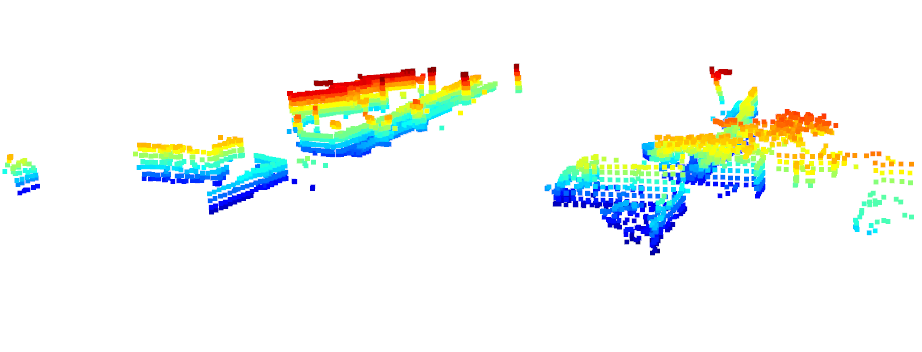}}&
    \hspace{-0.15in}{\setlength{\fboxrule}{2pt}\fcolorbox{red}{white}{\includegraphics[width=0.5\linewidth]{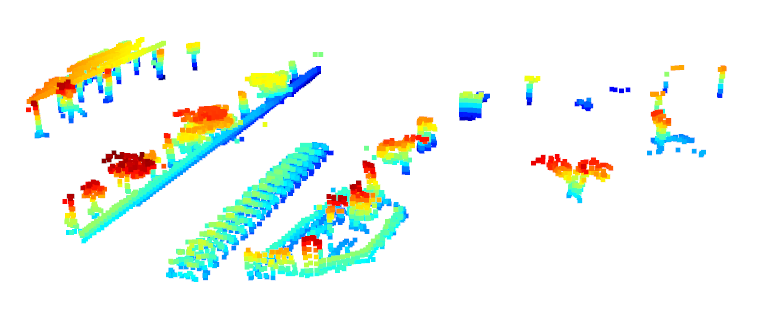}}} \\
    (e) &  (f) \\
    \fbox{\includegraphics[width=0.5\linewidth]{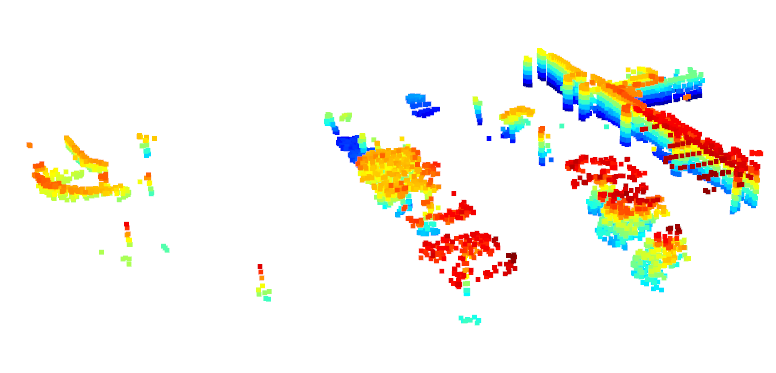}}&
    \hspace{-0.15in}{\setlength{\fboxrule}{2pt}\fcolorbox{red}{white}{\includegraphics[width=0.5\linewidth]{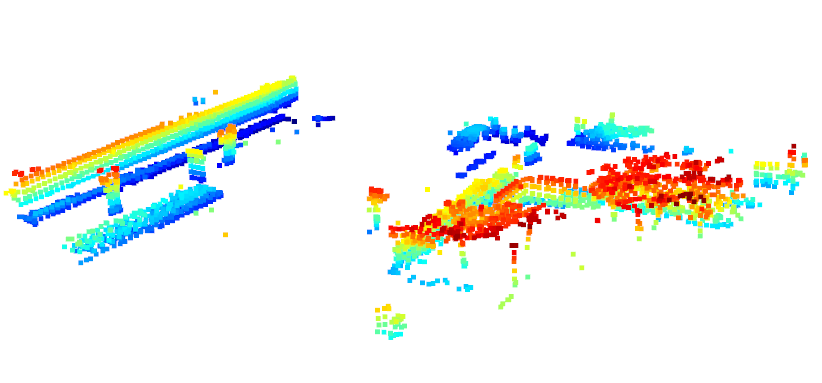}}} \\
    (g) &  (h) \\
    \end{tabular}
      
\caption{Examples of failed retrievals. A query (left) and its corresponding matched point cloud (right) from Oxford (a, b), NUS-U (c, d), NUS-R  (e, f), and NUS-B  (g, h). }
\label{fig:failure_oxford}
\end{figure}    


\subsubsection{Place Recognition: Oxford and NUS}
We tested our proposed \tda~framework on four real-world datasets, Oxford RobotCar and three from NUS, and compared the Recall-1\% values against two learning-based baselines, i.e., PointNetVlad and PCAN, and two handcrafted baselines, namely M2DP~\cite{he2016m2dp} and ScanContext (SC)~\cite{kim2018scan}. These are a few of the most popular baselines for place recognition. SC and M2DP are the closest to our approach, as they also do not employ deep learning techniques for finding descriptors. The results for PointNetVlad and PCAN are reported from their respective papers, whereas the same settings from SC and M2DP are used in our implementations. For SC, we used its official implementation from MATLAB\footnote{https://www.mathworks.com/help/vision/ref/scancontextdescriptor.html} while using the authors' MATLAB code for M2DP\footnote{https://github.com/LiHeUA/M2DP/tree/master}. The results are presented in Table \ref{tab:recall1p_comp}. This result demonstrates the strength of the \tda~framework in a large-scale place recognition application using point clouds.

\begin{figure}[ht!]
    \centering
\hspace{-0.1in}\includegraphics[width=0.63\linewidth]{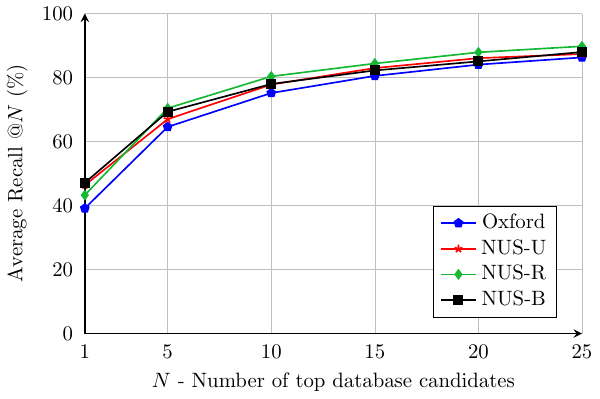}
    \caption{Change in the Recall @$N$ metric values for various values of $N$.}
    \label{fig:recallN_avg}
    \vspace{-15pt}
\end{figure}

\begin{figure*}[h]
\centering
\includegraphics[width=0.26\linewidth]{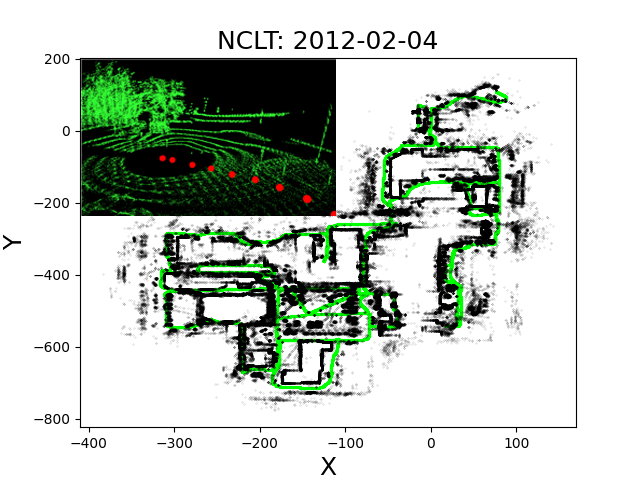}\hspace{-0.1in}\includegraphics[width=0.26\linewidth]{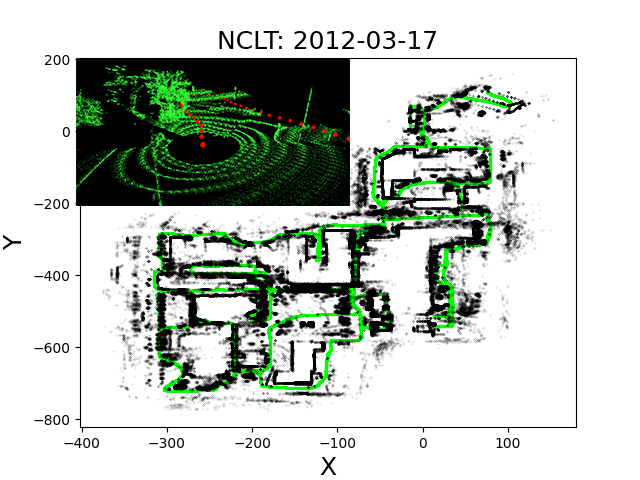}\hspace{-0.1in}\includegraphics[width=0.26\linewidth]{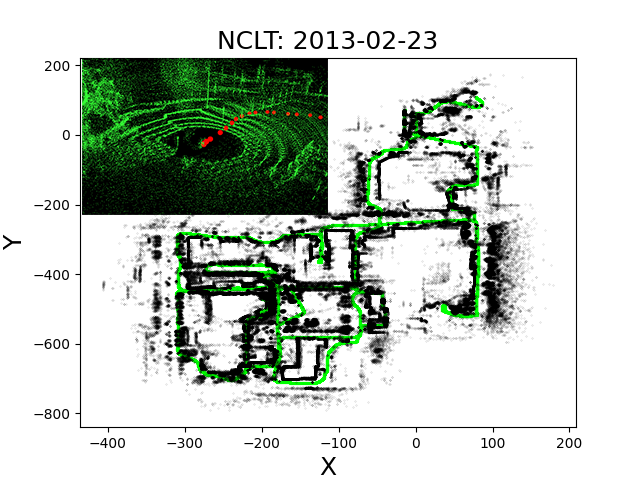}\hspace{-0.1in}\includegraphics[width=0.26\linewidth]{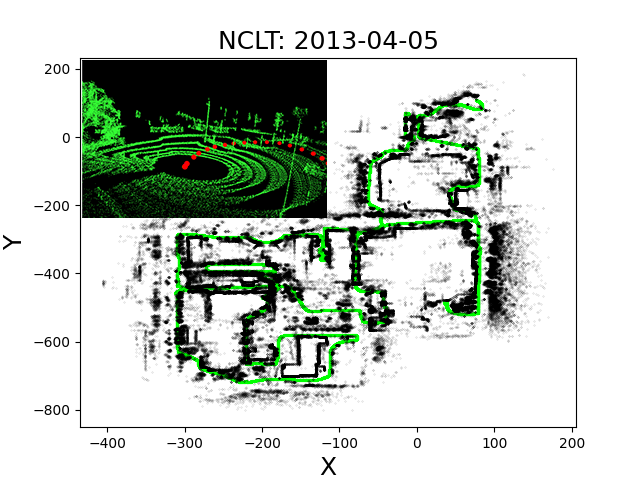}
\caption{The trajectories and sample scans for the four NCLT datasets (2012-02-04, 2012-03-17, 2013-02-23, 2013-04-05) are shown.}
\label{fig:NCLT_trajectories}
\end{figure*}

\begin{figure*}[ht!]
    \centering
    \includegraphics[trim={1cm 0cm 2cm 0cm}, clip, width=0.25\linewidth]{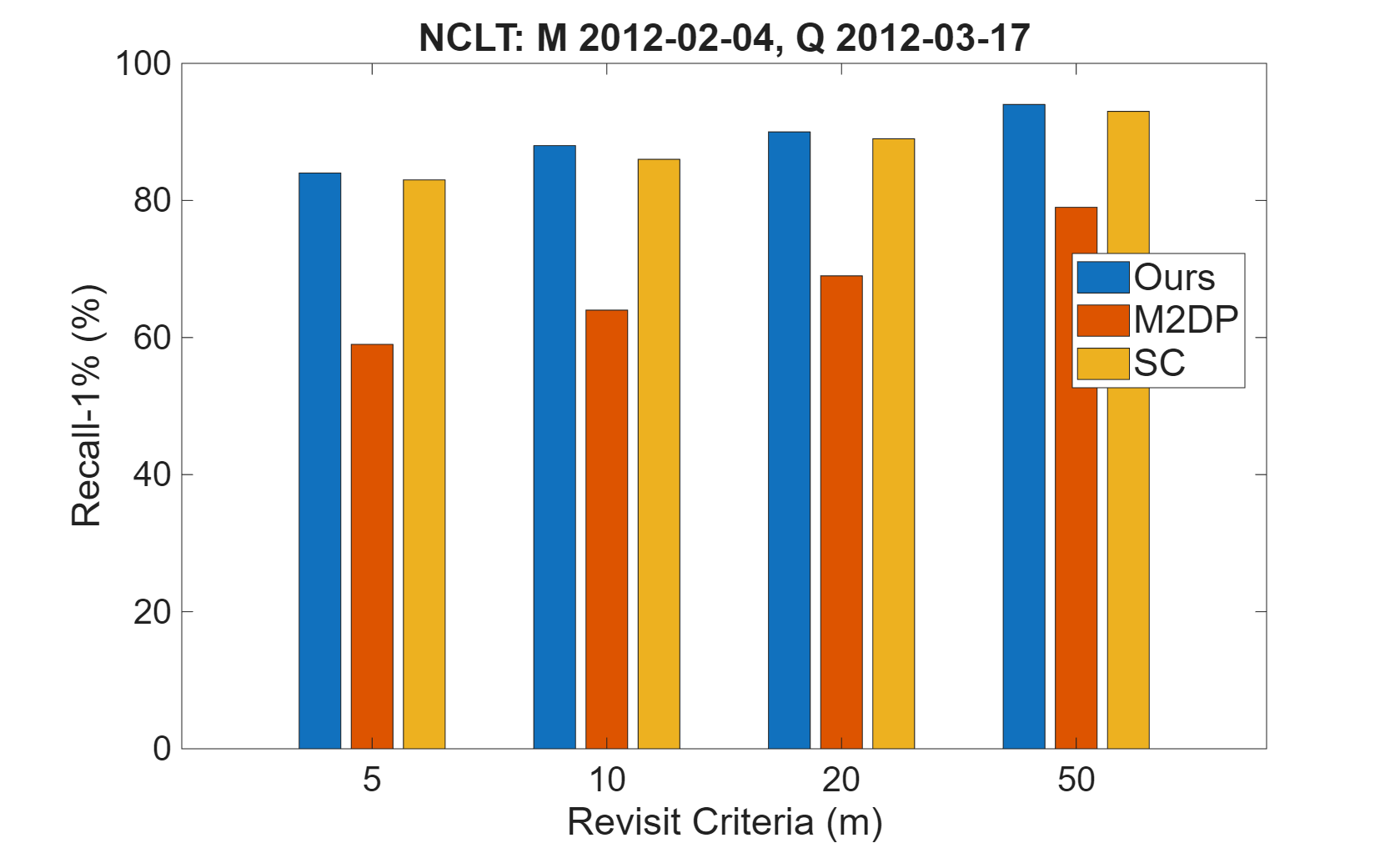}\includegraphics[trim={1cm 0cm 2cm 0cm}, clip, width=0.25\linewidth]{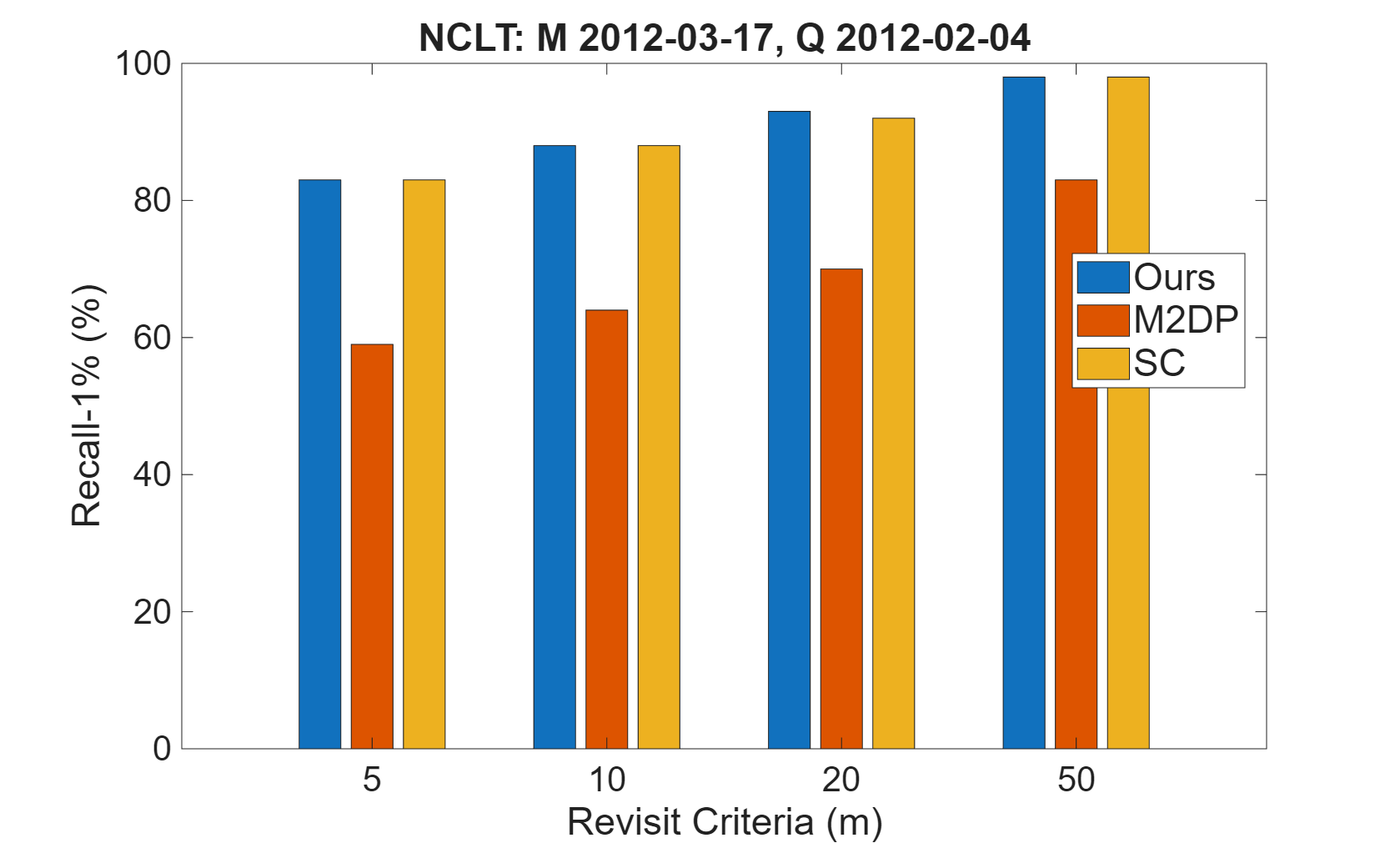}\includegraphics[trim={1cm 0cm 2cm 0cm}, clip, width=0.25\linewidth]{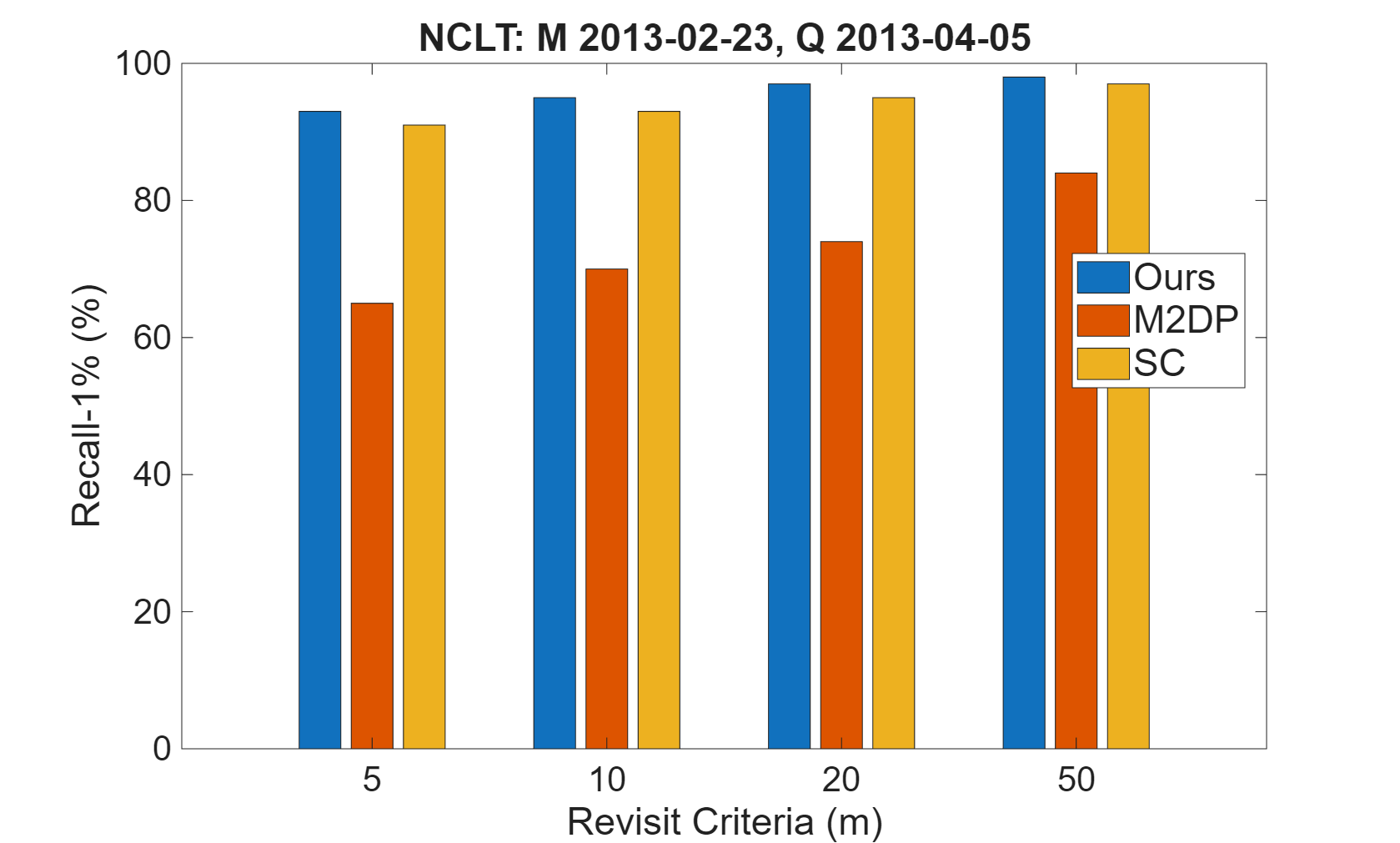}\includegraphics[trim={1cm 0cm 2cm 0cm}, clip, width=0.25\linewidth]{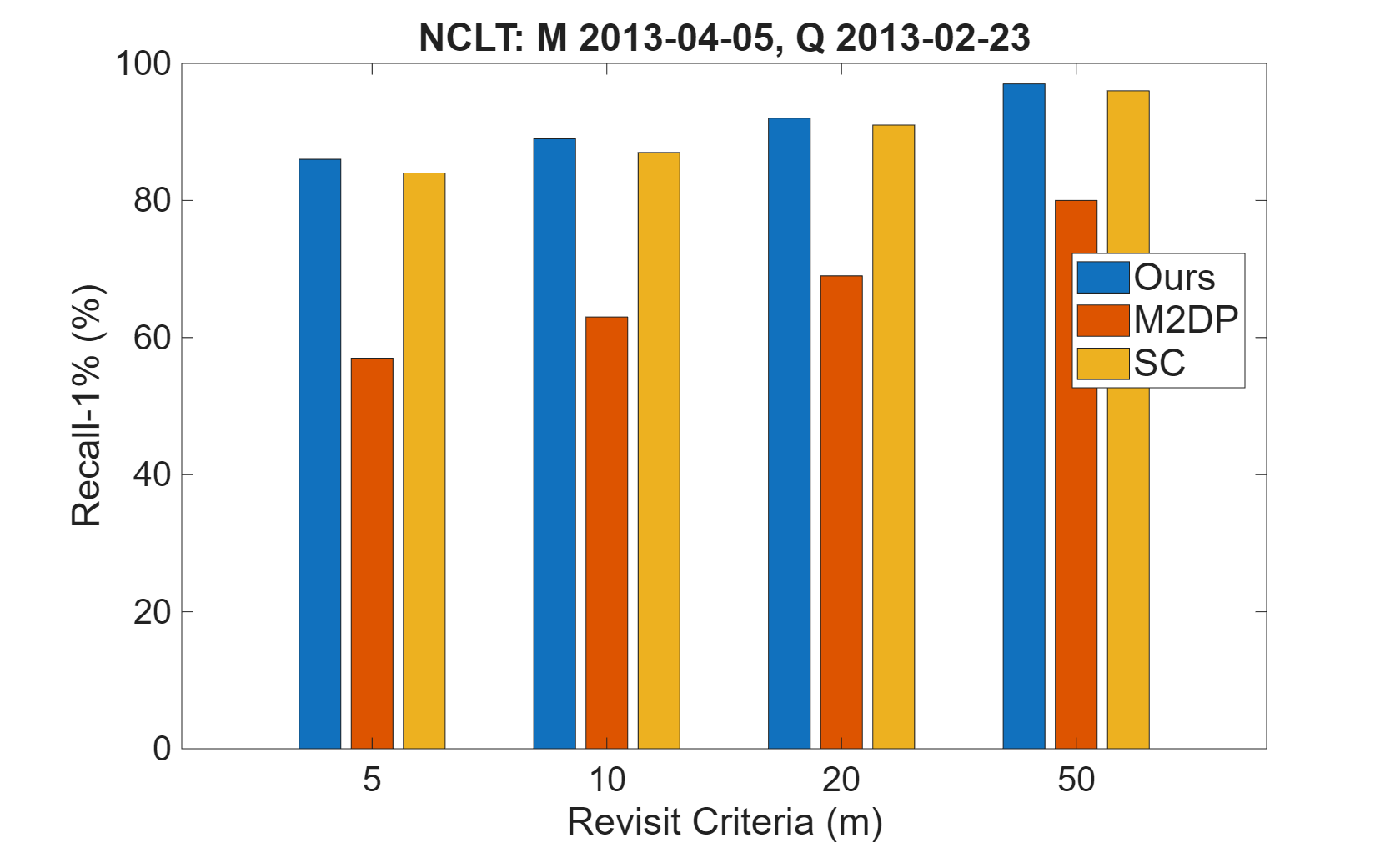}
    \caption{Recall-1\% metric comparison for place recognition on the NCLT dataset (case 1-4) with $b=1000$.}
    \label{figs:NCLT_recall}
\end{figure*}

Remember that, unlike the PointNetVlad and PCAN baselines used here, our proposed technique does not require extensive training and GPU computations. For example, it took \tda~just $325, 61, 47, \text{and } 47$ sec. for the datasets Oxford, NUS-U, NUS-R, and NUS-B, respectively, for pre-processing steps 1 and 2 in Algo. \ref{fig:alg}. Still, \tda~could outperform PointNetVlad by $15.3$, $10.3$,	$21.4$, $7.9$ percentages on Oxford, NUS-U, NUS-R, and NUS-B datasets, respectively. On the other hand, we outperformed PCAN by $11.8$, $3.8$, $10.5$, $6.4$ percentages on those datasets. Note that the query times are reported in Table \ref{tab:time_NCLT}. Overall, \tda~outperformed PointNetVlad and PCAN by $13.72$ and $8.12$ percentages across these four real-world large-scale datasets used as a standard for point cloud-based place recognition. 

When compared with SC and M2DP, we find that for the Oxford and NUS-U datasets, \tda~ achieved superior performance to these traditional baselines by a few percentage points. While SC and M2DP slightly outperformed \tda~in NUS-R and NUS-B datasets, our approach delivered superior results in two other important datasets, underscoring its overall robustness and competitiveness. 
Representative successful and failed retrievals for these datasets are shown in Figs. \ref{fig:queryandmatch_oxford} and \ref{fig:failure_oxford}, respectively.

We also investigate Recall @$N$ values for \tda~ on the Oxford and NUS datasets. The result is presented in Fig. \ref{fig:recallN_avg}. Recall @$N$ increased with the increase in $N$, a trend that was also observed for PointNetVlad~\cite{uy2018pointnetvlad}. 

\subsubsection{Place Recognition: NCLT}

The Recall-1\% results are presented in Fig.~\ref{figs:NCLT_recall}. In all these tests, we have comprehensively outperformed M2DP - the maximum difference being $29\%$ (case 4, 5m). We have also outperformed SC in all but one variation when \tda~and SC achieved the same Recall-1\% value (case 2, 50m). The maximum difference with SC is $2\%$. 
Summarizing all experiments, the proposed \tda~achieved better performance in 17 out of the total test cases, while M2DP and ScanContext led in only one case each (excluding the one tie between SC and \tda), demonstrating clear overall superiority.

\noindent
\textbf{Query Processing Time. }The average query processing time to retrieve the best match for the tested approaches is listed in Table \ref{tab:time_NCLT}. The NUS variants and Oxford did not introduce meaningful variance in run time due to the standardized $4096$-sized scans, and therefore, are listed together in Table \ref{tab:time_NCLT}. This result shows that the SC and \tda's execution times are competitive, while M2DP is the fastest. 


\begin{table}[ht!]
    \centering
    \begin{tabular}{cccc}
    \toprule
       Dataset \textbackslash~Method & Ours & SC & M2DP \\
        \midrule
        Oxford/NUS & 0.2 & 0.1 & 0.03\\
        NCLT & 0.1  & 0.1 & 0.08\\
         \bottomrule
    \end{tabular}
    \caption{Average run time (sec) per test scan.}
    \label{tab:time_NCLT}
\end{table}

We observed that different values of the ATOL descriptor length $b$ are more suitable for different datasets. Currently, finalizing this value before query evaluation is a manual process. Exploring an automated topological method to determine the optimal (or near-optimal) length $b$ based on persistence in a future work would be valuable.